\pdfoutput=1

\documentclass[11pt]{article}

\usepackage{ACL}

\usepackage{times}
\usepackage{latexsym}

\usepackage[T1]{fontenc}

\usepackage[utf8]{inputenc}

\usepackage{microtype}

\usepackage{inconsolata}

\usepackage{amsmath}
\usepackage{amsfonts}
\usepackage{graphicx}
\usepackage{subcaption}
\usepackage{tikz}
\usepackage{multirow}
\usepackage{float}
\usepackage{booktabs}
\usepackage{tcolorbox}
\usepackage{float}
\usepackage{multicol}
\usepackage{algorithm}
\usepackage{algpseudocode}
\usepackage{soul}
\usepackage{array}
\usepackage{arydshln}
\usepackage{bbm}
\usetikzlibrary{bayesnet}
\usepackage{xcolor}
\usepackage{colortbl}
\usepackage[edges]{forest}
\usepackage{xcolor}
\usepackage[utf8]{inputenc}
\usepackage{tcolorbox}
\usepackage{xcolor}
\usepackage{pifont}

\title{Bridging Compositional and Distributional Semantics: \\ A Survey on Latent Semantic Geometry via AutoEncoder}

\author{Yingji Zhang$^{1\dagger}$, ~ Danilo S. Carvalho$^{1,3}$, ~ Andr\'{e} Freitas$^{1,2,3}$ \\
$^{1}$ Department of Computer Science, University of Manchester, UK\\
$^{2}$ Idiap Research Institute, Switzerland\\
$^{3}$ Cancer Biomarker Centre, CRUK Manchester Institute, UK\\
\texttt{$^{1}$\{firstname.lastname\}@[postgrad.]$^{\dagger}$manchester.ac.uk}
\\ \texttt{$^{3}$\{firstname.lastname\}@idiap.ch}}
\begin{document}
\maketitle
\begin{abstract}
Integrating compositional and symbolic properties into distributional semantic spaces can enhance the interpretability, controllability, composition, and generalisation capabilities of Transformer Language Models. In this survey, we offer a novel perspective on latent space geometry through the lens of compositional semantics, a direction we refer to as \textit{semantic representation learning}. This direction enables a bridge between symbolic and distributional semantics, thus improving semantic and geometrical interpretability and localised control. We review and compare three mainstream autoencoder architectures: Variational AutoEncoder (VAE), Vector Quantised VAE (VQ-VAE), and Sparse AutoEncoder (SAE), and examine the distinctive latent geometries they induce in relation to semantic structure and interpretability.
\end{abstract}
\section{Introduction}

Over the past decade, language models have progressed from n-gram methods and RNNs to large Transformer-based architectures capable of human-level language understanding and generation ~\cite{vaswani2017attention,devlin2019bert,radford2019language}. 
However, this remarkable progress has come at a cost: we understand little about the internal semantic mechanisms these models use. They operate as “black boxes”, where the representations they learn are difficult to interpret and control. Specifically, \textbf{\textit{Interpretability:}} Neural networks discover useful features automatically, and their latent spaces can capture the underlying features in data in terms of geometry~\cite{Bengio2013RepresentationLA,alshammariunifying}. For Transformer LMs (TLMs), however, this latent space is high-dimensional, complex, with generative features for natural language poorly defined and geometrically entangled~\cite{scherlis2022polysemanticity}, thereby lacking \textit{semantic and geometrical} interpretation. \textbf{\textit{Controllability:}} Without a disentangled latent space, \textit{fine-grained, quasi-symbolic, localised} generative control is difficult \cite{yu2026latent}.

A promising path toward interpretability and control lies in understanding and shaping the latent space geometry. An interpretable latent geometry exhibits meaningful clustering and separation aligned with underlying generative features, enabling localised control via targeted manipulations. The autoencoder architectures, such as VAE~\cite{kingma2013auto}, VQ-VAE~\cite{van2017neural}, and SAE~\cite{shu2025survey}, offers a compelling framework for this. Their encoder–bottleneck–decoder structure encode inputs into a latent space, encouraging the capture of salient generative factors. When combined with an autoregressive TLM, these latent variables can be explicitly manipulated to guide generation. Crucially, we can intentionally shape this geometry for greater interpretability and controllability. E.g., latent generative factors can be potentially aligned with human-interpretable features, such as sentiment, semantic role, syntax, and reasoning behaviours, among others. 

Overall, this survey focuses on \textit{semantic representation learning} to render language representations or models more semantically and geometrically interpretable, and to enable localised, quasi-symbolic, compositional control through deliberate shaping of their latent space geometry. Our objective is to move toward TLMs whose internal semantic representations can be systematically interpreted, precisely shaped, and reliably directed. This type of representation can provide the foundation to shorten the gap between deep latent semantics and symbolic, linguistic representations, integrating the flexibility of distributional-neural models with the properties of linguistically grounded representations, facilitating both interpretability and generative control.




\begin{figure*}[tb]
  \includegraphics[width=\linewidth]{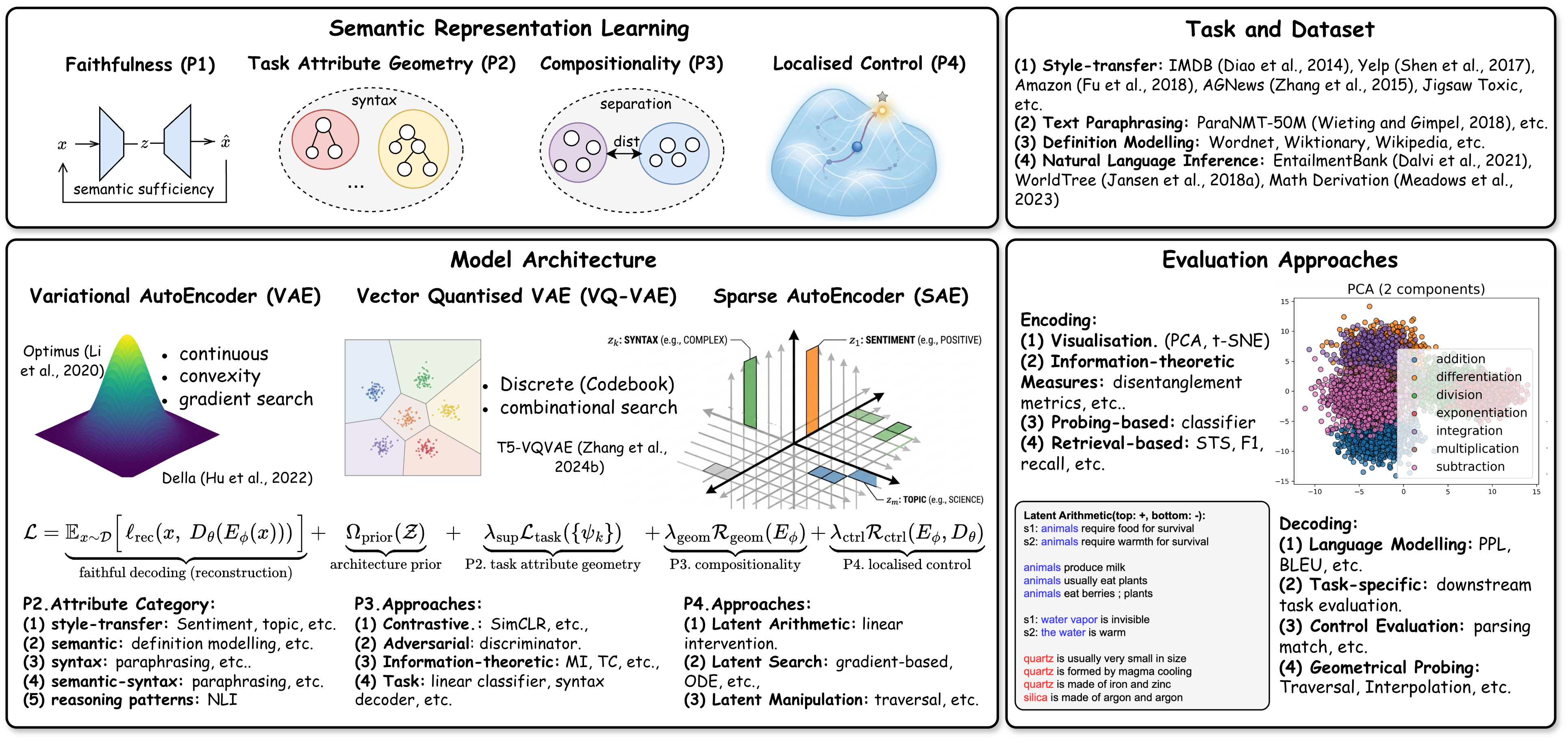}
  \caption{This survey focuses on the semantic and geometrical interpretability and fine-grained, localised, quasi-symbolic, compositional generative control of Transformer Language Models (TLMs) by shaping the latent space geometry within the AutoEncoder framework: VAE, VQVAE, and SAE.}
  \label{fig:overview}
\end{figure*}
\section{Semantic Representation Learning} \label{sec:related}

In this section, we introduce a unified perspective on semantic representation learning, grounded in the view of linguistic semantics as geometric structures in latent space. Rather than proposing a strict taxonomy, we synthesise insights from prior literature to identify four core properties that characterise desirable semantic representations.


Formally, let $\mathcal{X}$ be a space of linguistic expressions over a vocabulary $\mathcal{V}$. Let $\mathcal{F}=\{f_1,\ldots,f_K\}$ denote a set of \textit{semantic factors} (e.g., topic, sentiment, tense, semantic roles, reasoning rules), where each $f_k$ takes values in a domain $\mathcal{A}_k$ (finite or continuous). A \textit{semantic representation} consists of an encoder-decoder pair
$
E_\phi:\mathcal{X}\to\mathcal{Z},\qquad D_\theta:\mathcal{Z}\to \Delta(\mathcal{X}),
$
with latent space $\mathcal{Z}\subseteq\mathbb{R}^d$ (continuous, discrete, or mixed), such that the tuple $(E_\phi,D_\theta,\mathcal{Z})$ satisfies the following properties:

\textit{\textbf{P1.~Faithfulness (semantic sufficiency).}}
There exist readout maps $\{\psi_k:\mathcal{Z}\to\mathcal{A}_k\}_{k=1}^K$ (preferably linear or low-complexity\footnote{For instance, we can apply a ``linear'' classifier $\psi_k$ to identify factor $k$, where higher classification performance indicates that this factor can be encoded in the latent space \cite{zhang2025learningdisentanglelatentreasoning}.}) for which
$
\psi_k\!\left(E_\phi(x)\right) \approx f_k(x)\quad\text{for each }k,
$
and decoding preserves semantics in expectation:
\[
\mathbb{E}_{\tilde x\sim D_\theta(E_\phi(x))}\!\left[\mathbf{1}\{f_k(\tilde x)=f_k(x)\}\right]\ge 1-\epsilon_{\text{rec}},\ \ \forall k.
\]
This property serves as the foundational requirement, 
ensuring that semantic factors are sufficiently encoded and preserved, upon which higher-level properties (P2–P4) can be meaningfully defined.

\textit{\textbf{P2.~Task Attribute Geometry.}} For a given task, the latent space should be organised into task-relevant attribute spaces, where each semantic factor corresponds to a structured and interpretable subspace. Specifically, there is a decomposition of $\mathcal{Z}$ into factor-specific subspaces (e.g., sentiment-residual):
$
\mathcal{Z}=\bigoplus_{k=1}^K \mathcal{U}_k \oplus \mathcal{U}_{\mathrm{res}},
$
such that variation in $f_k$ is confined to $\mathcal{U}_k$ (up to tolerance $\epsilon_{\text{leak}}$). Writing $z=z_\parallel^{(k)}+z_\perp^{(k)}$ with $z_\parallel^{(k)}\in\mathcal{U}_k$ (projection of $z$ on $\mathcal{U}_k$) and $z_\perp^{(k)}\in \bigoplus_{j\ne k}\mathcal{U}_j\oplus\mathcal{U}_{\mathrm{res}}$, the readout $\psi_k$ depends only on $z_\parallel^{(k)}$. This property yields meaningful directions, distances, and clusters aligned to semantic features.



\textit{\textbf{P3.~Compositionality.}} 
The latent representation supports compositional structure over semantic attributes, such that factor-wise variations can be consistently combined and manipulated. This property is typically facilitated by learning strategies that promote factor-wise separability. Formally, there exists a latent operation $\oplus$ (such as addition) such that for any factor assignments $a,a'$:
$ E_\phi\!\big(\mathrm{compose}(x; f_k\!\leftarrow\!a \odot a')\big)\ \approx\ E_\phi\!\big(\mathrm{compose}(x; f_k\!\leftarrow\!a)\big)\ \oplus\ \Delta_k(a\!\to\!a')
$
with ``reusable'' composition vectors $\Delta_k(a\!\to\!a')\in\mathcal{U}_k$. For instance, learning sentiment steering vector from positive to negative by calculating the difference between two sets of instances, no matter how residual semantics varied \cite{mercatali-freitas-2021-disentangling-generative,shen2020educating}.

\textit{\textbf{P4.~Localised Control.}}
The model enables targeted manipulation of specific semantic attributes during generation. For any $k$ and target value $a'$, define an intervention operator
$
\mathcal{I}_k(z,a')\ :=\ z_\perp^{(k)}+\Gamma_k(a'), \Gamma_k:\mathcal{A}_k\to\mathcal{U}_k,
$
such that decoding $x'\sim D_\theta(\mathcal{I}_k(E_\phi(x),a'))$ achieves
$ \Pr\!\left[f_k(x')=a'\right]\ge 1-\epsilon_{\text{set}}, \mathrm{dist}\!\left(f_j(x'),\,f_j(x)\right)\le \epsilon_{\text{hold}}\ \ \forall j\ne k, \nonumber
$
That is, the success rate of factor intervention is higher than a probability, while the distance of unchanged information before and after intervention should less than a small value $\epsilon_{\text{hold}}$. 

Building on the four properties outlined above, semantic representation learning aims to shape the latent space to enhance both semantic and geometric interpretability, as well as to enable fine-grained, localised control over generation. 

\section{Architecture: A Latent Geometry Perspective}
This section presents a brief overview of the principal autoencoder architectures.

\textbf{\textit{(1) Transformer VAEs.}} VAEs~\cite{kingma2013auto} learn continuous latent manifolds endowed with a known geometric prior, typically an isotropic Gaussian. The encoder defines a mapping from discrete text inputs onto a smooth, low-dimensional Riemannian manifold. The decoder then learns a reverse mapping, generating text conditioned on coordinates sampled from this manifold. Through the KL regularisation term, the Gaussian prior induces an approximately Euclidean latent geometry, promoting global smoothness and convexity. These properties facilitate gradient-based optimisation \cite{bonnet2024searchinglatentprogramspaces} and enable latent arithmetic \cite{zhang-etal-2024-graph} within the representation space.
Modern Transformer VAEs (e.g., Optimus~\cite{li2020optimus}, Della~\cite{hu-etal-2022-fuse}, AdaVAE~\cite{tu2022adavae}, LlamaVAE~\cite{zhang2023llamavae}, LangVAE~\cite{carvalho2025langvae}) combine this latent geometry with pretrained encoders and decoders. The resulting latent space provides an abstract representation of natural language, enabling more global control over generation. While non-Euclidean latent geometries (e.g., hyperbolic~\cite{cho2023hyperbolic} or hyperspherical~\cite{davidson2018hyperspherical}) have been explored, most models retain Gaussian priors due to its mathematical tractability, ease of optimisation.

\textbf{\textit{(2) Vector Quantised VAEs.}}
Vector Quantised VAEs (VQ-VAEs)~\cite{van2017neural} replace the continuous latent manifold with a discrete latent geometry. The model learns a finite set of prototype vectors (the codebook), which can be interpreted as discrete points partitioning the representation space. The encoder performs a nearest-neighbour projection, mapping each input to its closest codebook vector under a chosen distance metric (typically Euclidean), or alternatively employs the Gumbel–Softmax relaxation to approximate discrete latent distributions in a differentiable manner. This induces a piecewise-constant latent space, where the geometry is segmented into Voronoi regions around codebook entries. Owing to this discrete structure, the latent space can be explored using combinatorial search strategies, such as greedy or beam search, to probe and control generation \cite{hosking-etal-2022-hierarchical}.

\textbf{\textit{(3) Sparse AutoEncoder.}}
In contrast to VAEs and VQ-VAEs, which primarily focus on learning compact latent representations, Sparse Autoencoders (SAEs) map inputs into higher-dimensional latent spaces, where representations are expressed as sparse combinations of basis feature vectors. For each input, only a small subset of these basis vectors is activated, transforming entangled representations into more disentangled and interpretable components. This design mitigates the issue of polysemanticity in large language models~\cite{elhage2022superposition}, where individual neurons or dimensions encode multiple unrelated features.

\paragraph{Optimisation.}
Under specific optimisation, all architecture has the potential to learn the latent representation satisfied the four properties. Formally, given data $\mathcal{D}=\{x_i\}$, we minimise
\begin{align} \label{eq:obj}
    \mathcal{L} & =  \underbrace{\mathbb{E}_{x\sim\mathcal{D}} \big[\,\ell_{\text{rec}}\!\left(x,\ D_\theta(E_\phi(x))\right)\big]}_{\text{faithful decoding (reconstruction)}} \\ \nonumber
    & + \underbrace{\Omega_{\text{prior}}(\mathcal{Z})}_{\text{architecture prior}} + \underbrace{\lambda_{\text{sup}} \mathcal{L}_{\text{task}}(\{\psi_k\})}_{\text{P2. task attribute geometry}} \\  \nonumber
    & + \underbrace{\lambda_{\text{geom}}\mathcal{R}_{\text{geom}}(E_\phi)}_{\substack{\text{P3. compositionality}}} + \underbrace{\lambda_{\text{ctrl}} \mathcal{R}_{\text{ctrl}}(E_\phi,D_\theta)}_{\substack{\text{P4. localised control}}} \nonumber
\end{align}
Here $\Omega_{\text{prior}}$ instantiates the architectural choice (e.g., dual-encoder \cite{zhang-etal-2024-graph}); $\mathcal{L}_{\text{task}}$ is a supervised objective to ensure that specific semantic attributes are explicitly encoded in the latent representation; $\mathcal{R}_{\text{geom}}$ is geometrical constraint to promote subspace separation (e.g., contrastive learning \cite{zhang2024truthx}, KL for VAE, sparsity for SAE); and $\mathcal{R}_{\text{ctrl}}$ enforces that $\mathcal{I}_k$ changes only factor $k$ (e.g., gradient searching \cite{bonnet2024searchinglatentprogramspaces}).

\begin{table}[h!]
\begin{tcolorbox}[
    colback=blue!5!white,    
    colframe=black,         
    width=\linewidth,       
    boxrule=0.5pt,          
    sharp corners,          
    fonttitle=\bfseries,    
    coltitle=black,         
    before skip=5pt,       
    after skip=5pt,         
    boxsep=0pt,
    left=5pt, right=5pt, top=5pt, bottom=5pt
]
\noindent \textbf{\textcolor{black}{Summary and Limitations.}}
\small
VAEs, VQ-VAEs, and SAEs exhibit distinct trade-offs in latent geometry. VAEs provide smooth, continuous spaces that can learn abstract-level representation and support interpolation and gradient-based optimisation. VQ-VAEs introduce discrete, clustered latent structures that improve interpretability, yet restrict compositionality and make optimisation less flexible. SAEs promote sparse, axis-aligned features that enable strong interpretability and localised control, though their linear structure may limit modelling of complex semantics. Overall, these approaches reflect a fundamental trade-off between continuity, discreteness, and interpretability in latent semantic geometry. SAE is unsupervised, which may be mis-aligned with task-specific feature.
\end{tcolorbox}
\end{table}

\begin{table*}[h!]
\centering
\resizebox{16cm}{!}{
\begin{tabular}{llllllllll}
\toprule
\textbf{Literature} & \textbf{Attributes} & \textbf{Architecture} & \textbf{Cont Opt.} & \textbf{Adv Opt.} & \textbf{Information Opt.} & \textbf{Task Opt.} & \textbf{Arithmetic} & \textbf{Search} & \textbf{Manipulate}\\ 
\hline \hline
\hline
\citet{https://doi.org/10.48550/arxiv.1703.00955} & sentiment & VAE & - & $\checkmark$ & - & $\checkmark$ (cls) & - & - & $\checkmark$ \\
\hline
\citet{fang2019implicit} & sentiment & VAE &- & $\checkmark$ & $\checkmark$ (posterior) & $\checkmark$ (cls) & $\checkmark$ & - & $\checkmark$ \\ \hline
\citet{john2019disentangled} & sentiment & VAE & - & $\checkmark$ &-& $\checkmark$ (BoW) & $\checkmark$ & - & - \\ \hline
\citet{mai-etal-2020-plug} & sentiment & AE & - & $\checkmark$ & -& $\checkmark$ (cls) & - & $\checkmark$ & - \\ \hline
\citet{shen2020educating} & sentiment & VAE & -& $\checkmark$ & - & - & $\checkmark$ & - & -  \\ \hline
\citet{rimsky-etal-2024-steering} & sentiment & SAE & - & - & - & - & $\checkmark$ & - & - \\ \hline
\citet{shi2021gtae} & sentiment & AE (graph) & - & - & - & - & - & - & $\checkmark$ \\ \hline
\citet{sha-lukasiewicz-2024-text} & sentiment & VAE & $\checkmark$ & - & - & - & - & - & $\checkmark$ \\ \hline
\citet{li-etal-2022-variational-autoencoder} & sentiment & VAE & - & - & $\checkmark$ (prior) & - & - & - & $\checkmark$ \\ \hline
\citet{xu2021vae} & sentiment; code-switch;... & VAE & $\checkmark$ & - & - & -& $\checkmark$ & - & - \\ \hline
\citet{hu2021causal} & sentiment, category & VAE & - & - & - & $\checkmark$ (cls) & - & - & $\checkmark$ \\
\hline
\citet{huang2020cycle} & sentiment; topic & AE & -& $\checkmark$ & - & - & - & - & $\checkmark$ \\ \hline
\citet{duan-etal-2020-pre} & sentiment; topic; length & VAE & - & $\checkmark$ & $\checkmark$ (prior) & - & - & - & $\checkmark$ \\ \hline
\citet{gu-etal-2022-distributional} & sentiment, topic, detox & VAE & - & - & $\checkmark$ (Euler distance) & $\checkmark$ (cls) & - & $\checkmark$ & - \\
\hline 
\citet{liu2020revision} & sentiment; topic & VAE & - & - & - & $\checkmark$ (BoW, cls) & - & $\checkmark$ & - \\ \hline
\citet{gu-etal-2023-controllable} & sentiment, topic, detox & AE+flow & - & - & $\checkmark$ (prior) & - & $\checkmark$ & $\checkmark$ & - \\ \hline
\citet{fang-etal-2022-controlled} & sentiment, topic, tense & VAE & -&- & $\checkmark$ (prior) & $\checkmark$ (cls) & - & - & $\checkmark$ \\ \hline
\citet{10.1145/3450946} & topic & VQ-VAE & - & - & - & $\checkmark$ & - & - \\ \hline
\citet{hu2024short} & topic & VQ-VAE & - & - & - & - & - & - & $\checkmark$ \\ \hline
\citet{yoo2024topic} & topic & VQ-VAE & - & - & - & - & - & - & $\checkmark$ \\ \hline
\citet{liu-etal-2023-composable} & negation, tense, formality & VAE & - & - & - & $\checkmark$ (cls) & - & $\checkmark$ & - \\ \hline
\citet{vasilakes-etal-2022-learning} & negation, uncertainty & VAE & - &  - & $\checkmark$ (MI) & $\checkmark$ (cls) & - & - & $\checkmark$ \\
\hline
\citet{he2025sae} & truthfulness,sentiment,... & SAE & $\checkmark$ & -& -& $\checkmark$ (cls) & $\checkmark$ & - & - \\ \hline
\citet{zhang2024truthx} &  truthfulness & AE* (dual) & $\checkmark$ & - & - & - & $\checkmark$ & - & $\checkmark$ \\ \hline
\citet{li2023inference} & truthfulness & SAE & - & - & - & - & $\checkmark$ & - & - \\ \hline
\citet{deng2025unveiling} & code-switch & SAE & - & - & - & - & $\checkmark$ & - & - \\  \hline
\citet{minder2025robustly} & chat-specific features, ... & SAE & - & - & - & - & $\checkmark$ & - & - \\ \hline
\citet{zhao-etal-2025-steering} & context-memory & SAE & $\checkmark$ & - & - & - & $\checkmark$ & - & - \\ \hline
\citet{yoshioka2022spoken} & fluent spoken & VAE (dual) & - & - & $\checkmark$ (posterior) & - & - & - & $\checkmark$ \\ \hline \hline
\citet{mercatali-freitas-2021-disentangling-generative} & syntactic factors & VAE (discrete) & - & - & $\checkmark$ (MI, TC) & - & $\checkmark$ & - & - \\ \hline
\citet{felhi2022towards} &  syntactic roles & VAE &  & - & - & - & - & - & $\checkmark$ \\ \hline
\citet{hosking-etal-2022-hierarchical} & hierarchical syntax & VQ-VAE & - & - & - & $\checkmark$(cls) & - & $\checkmark$ & - \\ \hline
\citet{huang2021generating} & semantic-syntax & AE (dual) & -& - & - & - & - & - & $\checkmark$ \\ \hline
\citet{zhang-etal-2019-syntax-infused} & semantic-syntax & VAE (dual) & - & - & $\checkmark$ (prior) & $\checkmark$ (treecoder) & - & -\\ \hline
\citet{bao2019generating} & semantic-syntax & VAE & - & $\checkmark$ & - & $\checkmark$ (cls,treecoder) & - & - & $\checkmark$ \\ \hline
\citet{chen-etal-2019-multi} & semantic-syntax & VAE (dual) & $\checkmark$ & - & $\checkmark$(prior,posterior) & - & - & - & - \\ \hline
\citet{huang-etal-2021-disentangling} & semantic-syntax & Transformer (dual) & - & $\checkmark$ & - & - &- & - & $\checkmark$ \\ \hline
\citet{felhi-etal-2022-exploiting} & semantic-syntax & VAE & - & - & - & - & - & - & $\checkmark$ \\ \hline
\citet{zhang-etal-2024-graph} & semantic-syntax & VAE (dual) & - & - & - & $\checkmark$ (cls) & - & - & $\checkmark$ \\ \hline
\citet{hosking-lapata-2021-factorising} & semantic-syntax & VQ-VAE & - & - & - & - & - & - & $\checkmark$ \\ \hline
\citet{carvalho2023learning} & semantic role & VAE & - & - & $\checkmark$ (prior) & - & $\checkmark$ & - & -  \\ \hline
\citet{zhang-etal-2024-learning} & semantic role & AE+flow & - & - & $\checkmark$ (prior) & - & $\checkmark$ & - & - \\ \hline
\citet{zhang2024formalsemanticgeometrytransformerbased} & semantic role & VAE & - & - & $\checkmark$ (prior) & - & $\checkmark$ & - & $\checkmark$ \\ \hline
\citet{roy-grangier-2019-unsupervised} & semantics (concept) & VQ-VAE* & - & - & - & - & - & - & $\checkmark$ \\ \hline
\citet{zhang-etal-2024-improving} & semantics (concept) & VQ-VAE & - & - & - & - &  $\checkmark$ & - & $\checkmark$ \\ \hline
\citet{garg2025crosslayerdiscreteconceptdiscovery} & semantics (concept) & VQ-VAE* & - & - & - & - & - & - & $\checkmark$ \\ \hline
\citet{jing2025sparse} & semantics & SAE & - & - & - & - & $\checkmark$ & - & - \\ \hline
\citet{zhang2024controllablenaturallanguageinference} & reasoning patterns & Transformer & - & - & - & - & - & - & $\checkmark$ \\ \hline
\citet{zhang2025learningdisentanglelatentreasoning} & reasoning patterns & VAE & - & - & $\checkmark$ (prior) & $\checkmark$ (cls) & - & - & $\checkmark$ \\ \hline
\citet{sanchez-etal-2023-hidden} & reasoning patterns & VAE & - & - & $\checkmark$ (prior) & - & - & - & $\checkmark$ \\ \hline
\citet{venhoff2025basemodelsknowreason} & reasoning patterns & SAE & - & - & - & - & $\checkmark$ & $\checkmark$ & - \\
\toprule
\end{tabular}
}
\caption{Existing work on \textbf{P2.~task attribute geometry} and \textbf{P3.~compositionality} optimisation spans contrastive (Cont), adversarial (Adv), information-theoretic (Information), and task-specific (Task) paradigms, while \textbf{P4.~localised control} is achieved via latent manipulation, arithmetic, or searching. The summarisation of dataset and evaluation metric used in those studies is provided in Table \ref{tab:bench_all}.}
\label{tab:learn_obj}
\end{table*}

\section{Semantic Representation Properties} \label{sec:latent_props}

This section outlines the major architectural trends for learning semantic representation spaces and their integration with language models.

\subsection{Task Attribute Geometry \textit{(P2)}}

Based on their distinct geometric properties with respect to task requirements, we categorise latent spaces into the following groups:

\textbf{\textit{(1) Style-transfer Attribute Space.}} Style-transfer attribute spaces refer to latent representations in which non-structural linguistic attributes (e.g., sentiment, topic) are disentangled from core semantic content. Within VAE-based frameworks, most studies focus on controlled generation for short texts, particularly sentiment transfer, where models aim to modify specific attributes (e.g., from positive to negative) while preserving the underlying meaning~\cite{li-etal-2022-variational-autoencoder, liu2020revision}. This paradigm has been extended to multi-attribute control, enabling the simultaneous manipulation of multiple stylistic dimensions~\cite{gu-etal-2023-controllable, gu-etal-2022-distributional, hu2021causal}.

Within VQ-VAE-based architectures, research has primarily focused on topic-level control. For example, Topic-VQVAE~\cite{yoo2024topic} extends this framework to document-level generation by learning discrete topic codes. In parallel, SAEs enable feature-level interventions in LLMs, supporting the steering of attributes such as sentiment, truthfulness, and political polarity~\cite{he2025sae}.

Building upon these attribute-based representations, we introduce four types of latent space geometries specifically designed to encode structural semantic features, such as syntax, semantics and reasoning logics.

\textbf{\textit{(2) Syntax Attribute Space.}}
We define syntax attribute spaces as latent representations where syntactic structure is encoded as independent generative factors, disentangled from lexical semantics. Most prior work focuses on modelling grammatical structures (e.g., subject–verb–object relations) to enable syntactically controlled generation, particularly in tasks such as style transfer and text paraphrasing, primarily within VAE-based frameworks~\cite{mercatali-freitas-2021-disentangling-generative, felhi2022towards}. In addition, HRQ-VAE \cite{hosking-etal-2022-hierarchical} models a hierarchical syntactic sketch space via residual quantisation, decomposing an input encoding into a sum of embeddings from multiple codebooks. Each codebook refines the residual of the previous one, enabling a coarse-to-fine representation of syntactic information.

\textbf{\textit{(3) Syntax-Semantic Attribute Space.}} 
Syntax-semantic attribute space refers to a class of latent representations in which both syntactic structure and lexical semantics are explicitly encoded in the latent space. This representation enables independent control over syntactic form and semantic content, facilitating structure-induced generation and compositional generalisation. Several studies have explored methods for learning such disentangled latent spaces, with most work focusing on text paraphrasing within VAEs~\cite{zhang-etal-2019-syntax-infused,hosking-lapata-2021-factorising}. In addition, \citet{zhang-etal-2024-graph} show that separating syntax and semantics via a graph-language dual-encoder architecture can mitigate the information bottleneck imposed by language encoders (e.g., BERT).

\textbf{\textit{(4) Semantic Attribute Space.}} 
semantic attribute spaces refer to latent representations structured around semantic role structure. Most work focuses on definition modelling and semantic role modelling, where latent variables capture role-specific meaning contributions~\cite{carvalho2023learning}. Recent approaches formalise this geometry using Argument Structure Theory, representing semantic role-word interactions as structured regions (e.g., convex cones) in latent space, enabling fine-grained control and interpretability~\cite{zhang-etal-2024-learning}. Discrete models (e.g., T5-VQVAE \cite{zhang-etal-2024-improving}) and SAEs \cite{jing2025sparse} further support controllable generation by introducing semantically interpretable latent features.

\textbf{\textit{(5) Reasoning Attribute Space.}} 
Reasoning attribute spaces refers to the latent representations that encode inferential structures or reasoning patterns as controllable factors. Recent studies focus on tasks such as rule-based natural language inference (NLI), where latent spaces capture logical relationships~\cite{zhang2024controllablenaturallanguageinference,zhang2025learningdisentanglelatentreasoning}. Related work also encodes symbolic graphs or reasoning trajectories (e.g., via random walks or sparse features), allowing models to reproduce structured reasoning behaviours during generation~\cite{sanchez-etal-2023-hidden, venhoff2025basemodelsknowreason}. For instance, \citet{venhoff2025basemodelsknowreason} use SAE to learn low-dimensional latent reasoning behaviours in LLMs, such as \textit{problem restatement}, \textit{subgoal setting}, and \textit{knowledge retrieval}. Applying these feature vectors during generation can guide base models to reproduce structured reasoning chains.

\begin{table}[h!]
\begin{tcolorbox}[
    colback=blue!5!white,    
    colframe=black,         
    width=\linewidth,       
    boxrule=0.5pt,          
    sharp corners,          
    fonttitle=\bfseries,    
    coltitle=black,         
    before skip=5pt,       
    after skip=5pt,         
    boxsep=0pt,
    left=5pt, right=5pt, top=5pt, bottom=5pt
]
\noindent \textbf{\textcolor{black}{Summary and Limitations.}}
\small
Overall, while these approaches mark a shift from unstructured latent representations toward geometrically organised spaces grounded in linguistic structures, structured and discourse-level representations remain underexplored. Moreover, compared to VAEs and VQ-VAEs, SAEs have received relatively limited attention in modeling structural representations. By encoding generative factors such as syntax, semantics, and reasoning, modern approaches enable more controllable, interpretable, and compositional language generation.
\end{tcolorbox}
\end{table}


\subsection{Compositionality \textit{(P3)}}
In this section, we review the principal optimisation strategies used to induce compositional structure in latent representations. Specifically, these methods aim to enforce factor-wise separability, thereby enabling consistent manipulation and combination of semantic attributes in the latent space.

\textbf{\textit{(1) Contrastive Learning.}} To encourage the latent space to capture contrastive features (e.g., positive-negative in style transfer), contrastive learning~\cite{sohn2016improved} is widely adopted as an effective objective. It operates on latent representations by promoting similarity between samples sharing the same underlying factor while separating those corresponding to different factors.
\[
\begin{aligned}
&\mathcal{L}(z, Z^+, Z^-) \\
&= - \log \frac{\sum_{z' \in Z^+} \exp\left(\mathrm{sim}(z, z') / \tau\right)}
{\sum_{z' \in Z^+ \cup Z^-} \exp\left(\mathrm{sim}(z, z') / \tau\right)} \\
\end{aligned}
\]
Here, $z$ denotes the anchor representation, $Z^+$ and $Z^-$ represent the sets of positive and negative samples, respectively. The function $\mathrm{sim}(\cdot,\cdot)$ denotes cosine similarity, and $\tau$ is a temperature parameter controlling the concentration of the distribution.

\textbf{\textit{(2) Adversarial Learning.}} Adversarial learning is employed to enforce invariance to non-target factors by removing undesired information from the latent representation. Typically, a discriminator is trained to predict non-target attributes from the latent code, while the encoder is optimised to prevent such predictions:
\[
\begin{aligned}
\min_{E} \max_{D} \mathcal{L}_{\text{adv}}(z, y) = \mathbb{E}_{z} \big[ \log D(y \mid z) \big],
\end{aligned}
\]
where $z = E(x)$ and $y$ denotes the non-target factor. This min–max objective encourages the latent space to be invariant to $y$, thereby promoting factor-wise disentanglement. 

However, adversarial objectives may introduce instability during training and can inadvertently remove information that is correlated with the target factor, leading to a trade-off between invariance and reconstruction fidelity.

\textbf{\textit{(3) Information-theoretic Optimisation.}} 
In contrast to contrastive and adversarial objectives, which operate at the sample level, information-theoretic optimisation regulates latent representations by explicitly controlling statistical dependencies at the distribution level. Representative approaches include: (i) Mutual Information (MI); (ii) Total Correlation (TC); and (iii) structured priors and posteriors, such as trainable priors~\cite{zhang2024formalsemanticgeometrytransformerbased} (i.e., Conditional VAE), that go beyond fixed prior assumptions. For instance, \citet{zhang2024formalsemanticgeometrytransformerbased} incorporate semantic role structure into the prior distribution of VAE using a BERT encoder. Under this formulation, the joint distribution is expressed as:
$
P_{\theta}(x,r,z)=\underbrace{P_{\theta}(x|z,r)}_{likelihood} \times \underbrace{P_{\theta}(z|r)}_{prior} \times P(r)
$
where the prior is conditioned on structured semantic role $r$. This design enables the latent space to encode structural information in a geometrically meaningful manner, thereby enhancing representational expressiveness. 

Nevertheless, information-theoretic optimisation typically incurs higher computational cost, as it requires estimating global statistical dependencies between variables, often through variational approximations or auxiliary networks.

\textbf{\textit{(4) Task Optimisation.}} Task-specific objectives are widely employed to induce latent separation by explicitly supervising representations with respect to target factors. Typical approaches include linear classifiers~\cite{he2025sae}, syntax-aware decoders~\cite{zhang-etal-2019-syntax-infused}, Bag-of-Words predictors~\cite{bao2019generating}, which impose structural or semantic constraints on the latent space. 

Nevertheless, this approach relies on the assumption that such information is already encoded in the latent space in a form that can be separated by a simple predictor. Consequently, deeper linguistic structures may not be adequately represented.

\begin{table}[h!]
\begin{tcolorbox}[
    colback=blue!5!white,    
    colframe=black,         
    width=\linewidth,       
    boxrule=0.5pt,          
    sharp corners,          
    fonttitle=\bfseries,    
    coltitle=black,         
    before skip=5pt,       
    after skip=5pt,         
    boxsep=0pt,
    left=5pt, right=5pt, top=5pt, bottom=5pt
]
\noindent \textbf{\textcolor{black}{Summary and Limitations.}}
\small
Overall, under different optimisation strategies, the latent space can represent specific geometry to target attributes. The structure of the latent space under different training objectives can be evaluated by visualising feature clusters and their separability. 
\end{tcolorbox}
\end{table}
\subsection{Localised Control \textit{(P4)}}
Having reviewed the principal techniques for learning latent representations, we now turn to the generation stage, focusing on methods for achieving localised control through the manipulation of latent representations. We categorise these approaches into three main classes.

\textbf{\textit{(1) Latent Manipulation.}} 
Latent manipulation refers to modifying latent representations, encompassing both continuous traversal~\cite{higgins2016beta, kim2018disentangling} and discrete interventions on semantically grounded variables (e.g., sentiment shifts).
For instance, we can perform a random walk, such as \textit{Ornstein-Uhlenbeck}, across the VAE latent space. If the latent space exhibits well-structured geometrical properties, the traversed examples are expected to display consistent patterns during generation.

\textbf{\textit{(2) Latent Arithmetic.}} Latent arithmetic provides a lightweight mechanism for localised control by exploiting linear structure in latent space. The underlying assumption is that semantic attributes are geometrically separable and approximately linear, allowing attribute manipulation via vector operations~\cite{Bengio2013RepresentationLA}.

Concretely, given an input $x$ with latent representation $z = E(x)$, attribute control can be achieved by decoding a perturbed code $D(z \pm v)$, where $v$ is a steering vector corresponding to the desired semantic shift~\cite{shen2020educating}. This paradigm has been extended to LLMs by intervening on residual activations~\cite{turner2023steering}, with methods such as CAA~\cite{rimsky-etal-2024-steering}, RePe~\cite{zou2023representation}, Top PC~\cite{im2025unified}, and ITI~\cite{li2023inference}, and SAE-based steering approaches~\cite{zhao-etal-2025-steering, he2025sae}.

However, this approach relies on the assumption that the latent space is approximately linear and convex, which limits its applicability.

\textbf{\textit{(3) Latent Searching.}} 
Latent searching refers to obtaining desired latent representations by iteratively optimising or navigating the latent space under a predefined objective. 

For example, \citet{bonnet2024searchinglatentprogramspaces} map input-output exemplars $\mathcal{D}=\{(x_i,y_i)\}_{i=1}^n$ into a continuous latent space and perform test-time adaptation by iteratively refining $z$ via gradient ascent:
$
z_{k+1} = z_k + \eta \nabla_z \sum_{i=1}^n \log p_\theta(y_i \mid x_i, z_k).
$
Alternatively, \citet{liu-etal-2023-composable} model latent navigation as an ODE driven by attribute-specific energy functions $E_i$ (e.g., attribute classifier), where a vector field guiding movement in latent space, yielding:
$
dz = \tfrac{1}{2}\beta(t)\sum_i \nabla_z E_i(a_i \mid z)\, dt.
$
For VQ-VAE, \citet{hosking-etal-2022-hierarchical} performs beam search over hierarchical latent syntax codes conditioned on semantic information, and uses the resulting latent sketch to control generation.

Although latent searching offers greater flexibility than latent manipulation and arithmetic, it typically requires higher computational cost and longer inference time.


\begin{table}[h!]
\begin{tcolorbox}[
    colback=blue!5!white,    
    colframe=black,         
    width=\linewidth,       
    boxrule=0.5pt,          
    sharp corners,          
    fonttitle=\bfseries,    
    coltitle=black,         
    before skip=5pt,       
    after skip=5pt,         
    boxsep=0pt,
    left=5pt, right=5pt, top=5pt, bottom=5pt
]
\noindent \textbf{\textcolor{black}{Summary and Limitations.}}
\small
Latent manipulation and latent arithmetic offer computationally efficient control mechanisms, but they rely on strong assumptions such as linearity and disentanglement. In contrast, latent search provides a more flexible and expressive framework, albeit with higher computational cost. VAEs with continuous latent space enable more efficient control and search than discrete latent spaces due to their smooth and differentiable geometry, which supports gradient-based optimisation. In contrast, discrete latent spaces require combinatorial or non-differentiable search, making optimisation over inherently discrete text representations particularly challenging.
\end{tcolorbox}
\end{table}

\section{Evaluation Approaches}
In this section, we review evaluation approaches from two perspectives: the latent space and the decoding process. These correspond to assessing four properties. We also summarise the benchmark datasets used in prior work in Tables~\ref{tab:bench_all} and \ref{tab:bench_all_1}.

\paragraph{Latent Space Evaluation.} Based on prior literature, latent representations can be evaluated along four key dimensions, each capturing different aspects of structure, information, and accessibility.

\textit{\textbf{(1) Visualisation.}} For VAE and VQ-VAE, dimensionality reduction techniques such as linear \textit{Principal Component Analysis (PCA)}, and non-linear \textit{t-SNE}~\cite{van2008visualizing} and \textit{UMAP}~\cite{mcinnes2018umap} are commonly employed to project high-dimensional latent representations into low-dimensional spaces. These visualisations provide qualitative evidence of clustering, separation, and continuity with respect to underlying semantic factors.

\textit{\textbf{(2) Information-theoretic Measures.}} Quantitative evaluation of latent information and structure can be performed using statistical measures such as MI and TC, as well as disentanglement metrics (as illustrated in Table \ref{tab:dis_eval}) including Mutual Information Gap (MIG), Modularity, Completeness, and Explicitness. Together, these metrics assess the extent to which individual latent dimensions capture distinct generative factors while remaining invariant to others \cite{carvalho2023learning}.

Compared to visualisation-based approaches, these metrics provide more rigorous and quantitative evaluation. However, they often rely on strong assumptions (e.g., factor independence or known ground-truth factors) and may not fully reflect the complexity of real-world linguistic representations.

\textit{\textbf{(3) Probing-based Evaluation.}} Linear prober (e.g., classifier) are trained on latent representations to predict target attributes (e.g., syntax depth in paraphrasing). High probing accuracy indicates that the corresponding factor is explicitly encoded and accessible in the latent space \cite{zhang-etal-2024-graph}.

\textit{\textbf{(4) Retrieval-based Evaluation.}} Latent representations can also be evaluated through semantic similarity and retrieval tasks. For example, performance on Semantic Textual Similarity (STS) benchmarks is typically measured using Pearson or Spearman correlation coefficients \cite{zhang2023llamavae}, while retrieval-specific metrics such as Recall and F1 score can be applied under appropriate experimental settings \cite{vasilakes-etal-2022-learning}. It can better evaluate complex semantic structure information in latent space. Its effectiveness, however, depends on the careful design of the target corpus and evaluation tasks.

\paragraph{Decoding Evaluation.}
While latent space evaluation focuses on representational structure, decoding evaluation assesses how such structure translates into generation quality and controllability.

\textit{\textbf{(1) Language Modelling Metrics.}} Standard metrics such as BLEU, ROUGE, and Perplexity (PPL) are used to evaluate fluency and reconstruction fidelity~\cite{li2020optimus}. These metrics primarily reflect P1 (faithfulness).

\textit{\textbf{(2) Task-specific Evaluation.}} Task-specific metrics evaluate performance with respect to downstream objectives. For example, in truthfulness steering, question answering performance can be measured using Exact Match (EM)~\cite{zhang2024truthx}. Compared to general language modelling metrics, these evaluations are more aligned with practical goals, but they are task-dependent and may lack comparability across settings.

\textit{\textbf{(3) Control Evaluation.}} The effectiveness of attribute control is typically measured using pretrained classifiers that assess whether generated outputs satisfy desired attributes~\cite{gu-etal-2022-distributional}. For structured tasks such as paraphrasing, controllability can also be quantified using metrics such as tree edit distance~\cite{zhang1989simple}. These approaches directly evaluate P4 (localised control).

\textit{\textbf{(4) Geometrical Probing.}} Geometrical probing evaluates controllability by applying latent operations such as traversal, interpolation, and arithmetic, and observing their effects on generated outputs~\cite{zhang-etal-2024-improving}. \citet{zhang-etal-2024-improving} further propose Interpolation Smoothness metric to quantify their evaluation. This provides intuitive evidence of latent structure and compositionality.



\begin{table}[h!]
\begin{tcolorbox}[
    colback=blue!5!white,    
    colframe=black,         
    width=\linewidth,       
    boxrule=0.5pt,          
    sharp corners,          
    fonttitle=\bfseries,    
    coltitle=black,         
    before skip=5pt,       
    after skip=5pt,         
    boxsep=0pt,
    left=5pt, right=5pt, top=5pt, bottom=5pt
]
\noindent \textbf{\textcolor{black}{Summary and Limitations.}}
\small
Overall, latent space evaluation assesses the structural quality of representations (e.g., via visualisation, probing, and information-theoretic metrics), while decoding evaluation measures generation quality and controllability. Together, they provide a complementary view of whether latent variables are both meaningful and useful.
However, current evaluation methods have notable limitations. Many metrics rely on simplifying assumptions (e.g., independence or linearity), qualitative analyses are subjective, and control evaluation often depends on biased pretrained classifiers. Moreover, improvements in latent disentanglement do not always translate to better generation performance. These issues highlight the need for more robust and standardised evaluation frameworks.
\end{tcolorbox}
\end{table}





\section{Challenges and Future Directions} \label{sec:concl}

In this section, we outline the potential challenges and future directions.


\textbf{\textit{(1) Latent Compositional Reasoning.}} Variational Autoencoders (VAEs) have been extensively investigated for generative drug design, where they are used to encode and decode chemical structures, thereby enabling the synthesis of novel molecular candidates \cite{ochiai2023variational}. In the neuro-symbolic domain, latent representations have also shown promise in facilitating the discovery of mathematical formulas \cite{li2026gensr}.

Nevertheless, the application of VAEs to scientific NLI, such as biomedical question answering that require multi-step entailment reasoning \cite{jullien-etal-2023-semeval}, remains relatively underexplored. This gap motivates the following research question: \textit{\textbf{RQ:} How can prior knowledge, reasoning structures, and logical constraints be efficiently encoded into latent representations to enhance NLI models?} Answering this question allows us to improve reasoning interpretability and safety.

\textbf{\textit{(2) Compositional Diffusion LMs.}}
diffusion-based language models provide a promising paradigm for capturing compositional semantics \cite{li2025survey}. In particular, syntactic structure can be explicitly integrated into the design of the diffusion process, such as through structured noise schedules that progressively corrupt and reconstruct representations in alignment with the hierarchical organization of language. This motivates the following research question: \textit{\textbf{RQ:} How can compositional language modelling be systematically achieved through the incorporation of diffusion processes?}

For instance, the model may first establish a coarse syntactic backbone (e.g., subject–verb–object structure), and then incrementally recover phrases, modifiers, and finer-grained lexical details. Under this paradigm, diffusion time steps can be aligned with the stages of language generation (from random noise $\rightarrow$ syntactic structure $\rightarrow$ semantic content $\rightarrow$ phrases $\rightarrow$ complete sentences). Such alignment introduces an inductive bias toward linguistic structure, thereby enhancing grammatical consistency, improving long-range coherence in sentence generation, and increasing controllability over the generation process.


\newpage

\section*{Limitations}
This survey primarily focuses on autoencoder-based frameworks (VAE, VQ-VAE, and Sparse Autoencoders) for shaping latent semantic geometry, and therefore does not comprehensively cover alternative interpretability or control paradigms outside this design space. In addition, the surveyed studies are predominantly based on English-language datasets, limiting insights into multilingual or low-resource settings. Finally, given the rapid evolution of language model interpretability research, some recent developments may not be fully captured.

\bibliography{tacl2021}
\bibliographystyle{acl_natbib}

\appendix
\section{Architecture Details} \label{sec:app1}

\paragraph{VAE.}
VAE comprises two primary components: an encoder and a decoder. The encoder maps an input \( x \) to a distribution over a latent space, typically parameterised as a Gaussian, while the decoder reconstructs the input by sampling from this latent distribution. Training optimises a variational objective consisting of a reconstruction term and a regularisation term that enforces proximity between the approximate posterior and a predefined prior over the latent variables, thereby encouraging a smooth and structured latent space:
\begin{align}
\mathcal{L}_{\text{VAE}}(x; \theta, \phi)
&= \mathbb{E}_{z \sim q_{\phi}(z|x)} \left[ \log p_{\theta}(x|z) \right] \nonumber \\
&\quad - \mathrm{KL}\!\left(q_{\phi}(z|x) \,\|\, p(z)\right).
\end{align}

A central geometric mechanism in such models is latent space injection, which determines how a sampled latent variable \( z \) influences the generative process. A representative instantiation is the Optimus architecture, which combines BERT~\cite{devlin2019bert} as the encoder with GPT-2~\cite{radford2019language} as the decoder. In this setup, BERT encodes the input sentence \( x \) into a fixed-length representation via the \texttt{[CLS]} token. This representation parameterises a Gaussian latent distribution \( \mathcal{N}(\mu, \Sigma) \), where both \( \mu \) and \( \Sigma \) are learned. A latent sample \( z \sim \mathcal{N}(\mu, \Sigma) \) is then injected into the decoder through different mechanisms, as summarised in Table~\ref{tab:latent_injection_formulas}.
\begin{table*}[h!]
\centering
\scriptsize
\begin{tabular}{p{7cm}p{8cm}}
\toprule
\textbf{Latent injection setup} & \textbf{Summarisation} \\ 
\hline \hline
$$\text{softmax}\left(\frac{Q [z;K]^T}{\sqrt{d}}\right) [z;V]$$ & \textbf{KV Memory}: LM decoder only rely on attached token $z$ which is more likely to be ignored by self-attention, even exacerbating KL vanishing~\cite{li2020optimus}. \\
\hline
$$\text{softmax}\left(\frac{Q [z + K]^T}{\sqrt{d}}\right) [z + V]$$ & \textbf{KV Addition}: Simple and global interventions in the sequence of attention of LM decoder, which can mitigate the KL vanishing without additional parameters~\cite{zhang-etal-2024-graph}. \\
\hline 
$$\text{softmax}\left(\frac{[Q+z] K^T}{\sqrt{d}}\right) V$$ & \textbf{Query Addition}: Integrating the latent representation $z$ into the decoder via query can deliver effective searching over Key and Value in decoders' parametric space~\cite{zhang2025learningdisentanglelatentreasoning,zhong2025understanding}. \\
\hline 
$$\hat{K} = \left(\sum_{i=1}^{r} W^i_k [K;1]\right) \circ \left(\sum_{i=1}^{r} W^i_z [z;1]\right)$$ & \textbf{KV Fusion}: Using tensor fuse operation~\cite{liu-etal-2018-efficient-low} to deeply fuse latent variables with hidden states in the LM decoder, but requiring additional training parameters~\cite{hu-etal-2022-fuse}. \\
\hline
$$h_{\text{cache}} = W_m z$$ & \textbf{KV Cache}: Flexibly integrating $z$ into the attention of various LM decoders with the cost of perturbing pretrained knowledge and more training parameters~\cite{carvalho2025langvae}. \\
\hline
$$P = \text{MLP}(z)$$ (where $P$ are $k$ soft prompt embeddings) & \textbf{Latent Prompt}: Seamlessly and flexibly integrating $z$ into LM decoders, but work on high-dimensional token embeddings. High dimensionality and complexity of the sequence space makes the search/optimisation difficult~\cite{zhou2024difflm}.\\
\toprule
\end{tabular}
\caption{Latent space injection methods in Transformer-based Language VAE.}
\label{tab:latent_injection_formulas}
\end{table*}

\paragraph{VQ-VAE.} The VQ-VAE extends the VAE framework by introducing discrete latent representations. It consists of three components: (i) an encoder that maps the input \( x \) to continuous latent vectors, (ii) a quantisation module that replaces each vector with its nearest neighbour in a learned codebook, and (iii) a decoder that reconstructs the input from these quantised representations. The training objective combines a reconstruction loss, a codebook loss that aligns encoder outputs with codebook embeddings, and a commitment loss that encourages encoder outputs to remain close to the selected embeddings:
\begin{align}
\mathcal{L}_{\text{VQ-VAE}}
&= \underbrace{\| x - \hat{x} \|_2^2}_{\text{reconstruction loss}}
+ \underbrace{\| \mathrm{sg}[z_e(x)] - e \|_2^2}_{\text{codebook loss}} \nonumber \\
&\quad + \underbrace{\beta \| z_e(x) - \mathrm{sg}[e] \|_2^2}_{\text{commitment loss}}.
\end{align}

\paragraph{SAE.} SAE consists of a linear encoder that projects the input into a higher-dimensional latent space and a linear decoder that reconstructs the input from this representation. To enforce sparsity, an additional regularisation term—commonly an \( \ell_1 \) penalty or a KL-divergence-based constraint—is applied to the latent activations. This encourages most neurons to remain inactive for any given input, thereby promoting the learning of meaningful and disentangled features while preventing trivial identity mappings. A comprehensive survey of SAE architectures and optimisation strategies is provided by \citet{shu2025survey}. The objective is given by:
\begin{align}
\mathcal{L}_{\text{SAE}} =
\underbrace{\| x - \hat{x} \|_2^2}_{\text{reconstruction loss}}
+ \underbrace{\lambda \| z \|_1}_{\text{sparsity penalty}}.
\end{align}

\section{Dataset and Benchmark}
In this section, we introduce the main dataset and task used for semantic representation learning. A summary of the datasets for each study is provided in Tables~\ref{tab:bench_all} and~\ref{tab:bench_all_1}.

\paragraph{Style Transfer.}
Style transfer datasets capture attribute-driven semantic variations, such as sentiment, topic, or tone, and are widely used in controlled text generation tasks. Representative corpora include Yelp~\cite{shen2017style}, Yahoo Answers~\cite{yang2017improved}, dSentences~\cite{discreteVAE}, IMDb~\cite{maas-etal-2011-learning}, AGNews~\cite{zhang2015character}, and Jigsaw. In addition, SAEs have been explored as a mechanism for steering truthfulness in LLMs. Benchmarks used to evaluate this capability include TruthfulQA~\cite{lin-etal-2022-truthfulqa}, Natural Questions~\cite{kwiatkowski-etal-2019-natural}, TriviaQA~\cite{joshi-etal-2017-triviaqa}, and FACTOR~\cite{muhlgay-etal-2024-generating}.

\paragraph{Text Paraphrasing.}
Paraphrasing datasets focus on semantic equivalence under surface variation, providing pairs of sentences that convey the same meaning with different lexical or syntactic realisations. Prominent resources include ParaNMT-50M~\cite{wieting-gimpel-2018-paranmt}, the Penn Treebank (PTB)~\cite{marcus-etal-1993-building}, SNLI~\cite{bowman-etal-2015-large}, and the Quora Question Pairs dataset\footnote{\url{https://www.kaggle.com/c/quora-question-pairs}}.

\paragraph{Definition Modelling.}
Definition modelling datasets capture mappings from lexical items to their corresponding conceptual descriptions, often following canonical patterns such as \textit{``X is a Y that Z''}. These resources are used to study the transformation between a word (or its embedding) and its textual definition~\cite{noraset2016definition,mickus-etal-2022-semeval}. Commonly used corpora include Wikipedia, WordNet, and Wiktionary.

\paragraph{Natural Language Inference.}
Datasets with structured explanations, such as atomic and conditional statements, encode multi-step reasoning in natural language and serve as key benchmarks for compositional inference. Representative resources include WorldTree~\cite{jansen2018worldtree}, EntailmentBank~\cite{dalvi2021explaining}, and Zebra~\cite{molfese-etal-2024-zebra}, all of which provide explicit reasoning chains that facilitate the study of compositional generalisation.

In addition, the Math Derivation dataset~\cite{meadows2023symbolic} introduces a highly controlled corpus for symbolic reasoning. In this setting, models are required to generate derived mathematical expressions given an input expression and a specified operation (e.g., differentiation, addition, or subtraction). To evaluate out-of-distribution (OOD) generalisation and assess whether models capture underlying reasoning patterns, the dataset defines five evaluation splits:

(1) EVAL: mathematical expressions following the training set's distribution (like $U + cos{(n)}$), (2) VAR: mathematical expressions composed of a different set of variables (like $U + cos{(beta)}$), (3) EASY: mathematical expressions with a lower number of variables (like $cos{(n)}$), (4) EQ: mathematical derivations with equality insertions (like $E=U+cos{(n)}$), (5) LEN:  mathematical derivations with a higher number of variables (like $U + cos{(n)}) + A + B$).

Moreover, SyllogisticNLI~\cite{valentino2025mitigatingcontenteffectsreasoning} introduces a controlled benchmark for syllogistic deductive reasoning, enabling the evaluation of logical inference under minimal confounding linguistic variation.

\begin{table*}[h!]
\centering
\resizebox{16cm}{!}{
\begin{tabular}{lp{9cm}p{2.1cm}p{4.4cm}}
\toprule
\textbf{Literature} & \textbf{Task, Dataset, and Benchmark} & \textbf{Encoding} & \textbf{Decoding} \\ 
\hline \hline
\citet{https://doi.org/10.48550/arxiv.1703.00955} & Sentiment Transfer: IMDB \cite{10.1145/2623330.2623758}, SST-full \cite{socher-etal-2013-recursive}, Lexicon \cite{wilson-etal-2005-recognizing} & - & Classifier Acc \\
\hline
\citet{fang2019implicit} & Sentiment Transfer: PTB \cite{marcus-etal-1993-building}, Yahoo, Yelp \cite{he2019lagging,yang2017improved} & KL, MI, AU & Classifier Acc, BLEU, PPL, RPPL, ... \\ \hline
\citet{john2019disentangled} & Sentiment Transfer: Yelp Review \cite{shen2017style}, Amazon Review \cite{fu2018style} & Cosine Sim \newline T-SNE & Classifier Acc, Word Overlap, PPL, ... \\ \hline
\citet{mai-etal-2020-plug} & Sentiment Transfer: WikiLarge \cite{zhang-lapata-2017-sentence}, Yelp Review \cite{shen2017style} & - & Classifier Acc, BLEU \\ \hline
\citet{shen2020educating} & Sentiment Transfer: Yelp, Yahoo \cite{yang2017improved} & Recall & Parser Acc, BLEU, PPL  \\ \hline
\citet{rimsky-etal-2024-steering} & LLM Steering: AI Risk and Sycophancy \cite{perez2023discovering} & PCA & QA match \\ \hline
\citet{shi2021gtae} & Sentiment Transfer: Yelp
review \cite{shen2017style}, Political Slant \cite{prabhumoye-etal-2018-style}  & - & WMD, PPL, Classifier Acc, BLEU \\ \hline
\citet{sha-lukasiewicz-2024-text} & Sentiment Transfer: Yelp
review \cite{shen2017style}, Amazon Review \cite{fu2018style} & MIG, T-SNE & Classifier Acc, PPL, \newline BLEU \\ \hline
\citet{li-etal-2022-variational-autoencoder} & Sentiment Transfer: EMPATHETIC, TACRED & T-SNE & Classifier Acc, WMD, BLEU \\ \hline
\citet{xu2021vae} & Sentiment/Code-switch/Formality Transfer: \newline Yelp, GYAFC \cite{rao-tetreault-2018-dear}, LinCE \cite{aguilar-etal-2020-lince} & - & Classifier Acc, PPL, \newline BLEU \\ \hline
\citet{hu2021causal} & Sentiment/Category Transfer: Yelp, BIOS \cite{de2019bias} & - & Classifier Acc, Fluency, Diversity \\
\hline
\citet{huang2020cycle} & Sentiment/Topic Transfer: Yelp, Yahoo QA & - &  Classifier Acc, BLEU, PPL, RPPL \\ \hline
\citet{duan-etal-2020-pre} & Sentiment/Topic/length Transfer: Yelp, News Titles \cite{fu2018style} & - & Classifier Acc, Diversity \cite{li-etal-2016-diversity} \\ \hline
\citet{gu-etal-2022-distributional} & Sentiment/Topic/Detox Transfer: IMDb \cite{maas-etal-2011-learning}, AGNews \cite{zhang2015character}, Jigsaw Toxic & PCA & Classifier Acc, Distinctness \cite{li-etal-2016-diversity} \\
\hline 
\citet{liu2020revision} & Sentiment/topic Transfer: Yelp, Amazon reviews \cite{he2016ups} & - & Classifier Acc, PPL, Word Overlap, BLEU \\ \hline
\citet{gu-etal-2023-controllable} & Sentiment/Topic/Detox Transfer: IMDb \cite{maas-etal-2011-learning}, AGNews \cite{zhang2015character}, Jigsaw Toxic & - & Classifier Acc, Distinctness \cite{li-etal-2016-diversity} \\ \hline
\citet{fang-etal-2022-controlled} & Sentiment/Topic/Tense Transfer: PTB \cite{marcus-etal-1993-building}, SNLI \cite{bowman-etal-2015-large}, Yahoo Answers \cite{xu2018spherical}, Yelp \cite{yang2017improved} & MI, AU & Classifier Acc, PPL, Distinctness \cite{li-etal-2016-diversity} \\ \hline
\citet{10.1145/3450946} & Topic Transfer: 20 Ng \cite{lang1995newsweeder}, Conala (https://conala-corpus.github.io/), & - & PPL, Topic Coherence, Topic Diversity, Topic Interpretability \\ \hline
\citet{hu2024short} & Topic Transfer: & \\ \hline
\citet{yoo2024topic} & Topic Transfer: 20 Ng \cite{lang1995newsweeder}, New York Times & T-SNE & Topic Coherence, Topic Diversity \\ \hline
\citet{liu-etal-2023-composable} & Sentiment/Negation/Tense/Formality Transfer: Yelp \cite{shen2017style}, Amazon Review \cite{fu2018style} & - & Classifier Acc, PPL, BLEU, CTC \cite{deng-etal-2021-compression} \\ \hline
\citet{vasilakes-etal-2022-learning} & Negation/Uncertainty Transfer: Yelp \cite{shen2017style}, Amazon Review \cite{fu2018style} &  T-SNE, Precision, recall, F1 & PPL, BLEU \\
\hline
\citet{he2025sae} & Truthfulness/Sentiment Steering: TruthGen \cite{fulay-etal-2024-relationship} & - & Steering Success Rate, Lexical Diversity, Entropy\\ \hline
\citet{zhang2024truthx} &  Truthfulness Steering: TruthfulQA \cite{lin-etal-2022-truthfulqa}, Natural Questions \cite{kwiatkowski-etal-2019-natural}, TriviaQA \cite{joshi-etal-2017-triviaqa}, FACTOR \cite{muhlgay-etal-2024-generating} & Kernel density estimate & QA match \\ \hline
\citet{li2023inference} & Truthfulness Steering: TruthfulQA \cite{lin-etal-2022-truthfulqa} & - & QA match, Cross Entropy, KL \\ \hline
\citet{deng2025unveiling} & Code-switch Steering: Flores-200 \cite{costa2022no} & - & Adversarial Language Identification, CrossLingual Continuation \\  \hline
\citet{minder2025robustly} & chat-specific features, ... & SAE  \\ \hline
\citet{zhao-etal-2025-steering} & Context-Memory Steering: NQSwap \cite{longpre-etal-2021-entity}, Macnoise \cite{hong-etal-2024-gullible} & - & QA match \\ \hline
\citet{yoshioka2022spoken} & Text-to-speech Synthesis: CSJ \cite{maekawa2000spontaneous}, KVJ & - & Classifier Acc, BLEU \\ \toprule
\end{tabular}
}
\caption{Dataset and evaluation metrics for task-specific attribute latent space model.}
\label{tab:bench_all}
\end{table*}

\begin{table*}[h!]
\centering
\resizebox{16cm}{!}{
\begin{tabular}{lp{8.5cm}p{3.5cm}p{3.5cm}}
\toprule
\textbf{Literature} & \textbf{Task, Dataset, and Benchmark} & \textbf{Encoding} & \textbf{Decoding} \\ 
\hline \hline
\multicolumn{4}{c}{\textbf{\textit{structural attribute space}}} \\ \hline
\citet{mercatali-freitas-2021-disentangling-generative} & Syntax Disentanglement: Yelp, dSentences & Z-min, Z-diff, MIG & Parser Acc \\ \hline
\citet{felhi2022towards} &  Syntax Disentanglement: SNLI \cite{bowman-etal-2015-large} & MIG & qualitative eval \\ \hline
\citet{hosking-etal-2022-hierarchical} & Question Paraphrasing: Paralex \cite{fader-etal-2013-paraphrase}, Quora Question Pairs (https://www.kaggle.com/c/quora-question-pairs), MSCOCO 2017 \cite{lin2014microsoft} & T-SNE & iBLEU, BLEU, Self-B \\ \hline
\citet{huang2021generating} & Paraphrasing: ParaNMT-50M \cite{wieting-gimpel-2018-paranmt}, Quora, PAN \cite{madnani-etal-2012-examining}, MRPC \cite{dolan-etal-2004-unsupervised} & - & BLEU, Parsing Match \\ \hline
\citet{zhang-etal-2019-syntax-infused} & Paraphrasing: PTB \cite{marcus-etal-1993-building}, Wikipedia \cite{gulordava-etal-2018-colorless} & - & Relevance, Readability, Diversity, PPL \\ \hline
\citet{bao2019generating} & Paraphrasing: PTB \cite{marcus-etal-1993-building}, Quora Question Pairs (https://www.kaggle.com/c/quora-question-pairs) & - & BLEU, PPL, RPPL, Tree Edit Distance \cite{zhang1989simple} \\ \hline
\citet{chen-etal-2019-multi} & Paraphrasing: ParaNMT-50M \cite{wieting-gimpel-2018-paranmt} & Pearson correlation (STS bench) & Tree Edit Distance, Constituent Parsing F1, POS Tagging \\ \hline
\citet{huang-etal-2021-disentangling} & Paraphrasing: ParaNMT-50M \cite{wieting-gimpel-2018-paranmt}, Quora Question Pairs (https://www.kaggle.com/c/quora-question-pairs) & Pearson correlation (STS bench), Classifier Acc (BShift, TreeDepth, TopConst), Cosine Sim & - \\ \hline
\citet{felhi-etal-2022-exploiting} & Semantic-syntax Disentanglement: ParaNMT-50M \cite{wieting-gimpel-2018-paranmt} & - & tree edit
distance, constituency tree (TMA2), ParaBART (consine sim), Meteor
score \\ \hline
\citet{zhang-etal-2024-graph} & Semantic-syntax Disentanglement: EntailmentBank \cite{dalvi2021explaining}, WorldTree \cite{jansen-etal-2018-worldtree}, Math Derivation \cite{meadows2023symbolic} & MI, KL, Wasserstein Distance, MSE, T-SNE & BLEU, BLEURT, Cosine Sim, PPL \\ \hline
\citet{hosking-lapata-2021-factorising} & Question Paraphrasing: Paralex \cite{fader-etal-2013-paraphrase}, Quora Question Pairs (https://www.kaggle.com/c/quora-question-pairs) & - & BLEU, Self-BLEU, iBLEU  \\ \hline
\citet{carvalho2023learning} & Semantic Disentanglement: Wordnet, Wiktionary, Wikipedia & T-SNE, UMAP, z-diff z-min-var, MIG Modularity, Explicitness, Disentanglement, Completeness, Informativeness  &  BLEU, PPL \\ \hline
\citet{zhang-etal-2024-learning} & Semantic Disentanglement: EntailmentBank \cite{dalvi2021explaining}, WorldTree \cite{jansen-etal-2018-worldtree} & T-SNE, PCA, Classifier Acc & Interpolation Smoothness \\ \hline
\citet{zhang2024formalsemanticgeometrytransformerbased} & Semantic Disentanglement: EntailmentBank \cite{dalvi2021explaining}, WorldTree \cite{jansen-etal-2018-worldtree} & T-SNE, PCA, Classifier Acc & Arithmetic Ratio \\ \hline
\citet{roy-grangier-2019-unsupervised} & Semantics (concept): WMT’17 English-German \cite{bojar-etal-2017-findings}, WMT-Newscrawl & & Classifier Acc, F1, BLEU, Regression \\ \hline
\citet{zhang-etal-2024-improving} & Semantics (concept): EntailmentBank \cite{dalvi2021explaining}, WorldTree \cite{jansen-etal-2018-worldtree}, Math Derivation \cite{meadows2023symbolic} & T-SNE & BLEU, PPL, Cosine, BLEURT, Interpolation Smoothness \\ \hline
\citet{garg2025crosslayerdiscreteconceptdiscovery} & Semantics (concept): ERASER \cite{pang-lee-2004-sentimental}, Jigsaw (https://kaggle.com/
competitions/jigsaw-toxic-comment-classification-challenge), AGNEWS & - & Fleiss’ Kappa, Average Confidence, Model Alignment Rate  \\ \hline
\citet{jing2025sparse} & Semantics: \cite{fromkin2017introduction}  & Feature Representation Confidence (FRC) & Feature Intervention Confidence (FIC)   \\ \hline
\citet{zhang2024controllablenaturallanguageinference} & Reasoning Disentanglement: EntailmentBank \cite{dalvi2021explaining} & PCA & BLEU, BLEURT, Cosine, LLM AutoEval \\ \hline
\citet{zhang2025learningdisentanglelatentreasoning} & Reasoning Disentanglement: EntailmentBank \cite{dalvi2021explaining}, Math Derivation \cite{meadows2023symbolic}, Syllogism & PCA & BLEU, Exact Match \\ \hline
\citet{sanchez-etal-2023-hidden} & Reasoning Graph: PTB \cite{marcus-etal-1993-building}, Yahoo, Yelp, ATOMIC \cite{sap2019atomic} & MI & BLEU, BERT Score \cite{Zhang*2020BERTScore:}, PPL \\ \hline
\citet{venhoff2025basemodelsknowreason} & reasoning patterns & SAE  \\
\toprule
\end{tabular}
}
\caption{Dataset and evaluation for structural attribute latent space model.}
\label{tab:bench_all_1}
\end{table*}

\begin{table*}[ht!]
\scriptsize
\centering
\renewcommand\arraystretch{1.1}
\begin{tabular}{p{3cm}p{2.5cm}p{9cm}}
\toprule
\textbf{Category} & \textbf{Metrics} & \textbf{Summarisation} \\ \hline \hline 
\multirow{3}{*}{Dimension-based} 
& $z_{min\_var}~error$ & Identify dimensions with the smallest variance across different samples for a given factor. \\
& \textit{Modularity} & Assess how exclusively a dimension is associated with a single factor. If MI for one factor dominates a dimension and others are close to zero, modularity is high. \\
& \textit{Disentanglement Score} & Use feature importances from a classifier to compute entropy for each dimension; lower entropy = better disentanglement. \\ \hline
\multirow{4}{*}{Factor-based} & 
\textit{Mutual Information Gap} & Measure how much more one latent dimension knows about a factor compared to others. A larger gap between the top two mutual information (MI) scores for a factor indicates stronger disentanglement.\\
& \textit{Completeness Score} & Similar to Disentanglement Score, but entropy is computed over factors rather than dimensions. \\
& \textit{Informativeness Score} & Use a classifier to predict factor labels from latent codes; score is based on prediction error (lower error = higher informativeness). \\
& \textit{Explicitness} & Determine how linearly separable the factor information is from the latent representation. Use ROC AUC to quantify separability.\\
\toprule
\end{tabular}
\caption{Disentanglement metrics quantify the feature-dimension alignment. An implementation in Python is provided by \cite{carvalho2025langvae}.} \label{tab:dis_eval}
\end{table*}

\section{Disentanglement Metrics} \label{app:dis}
Disentanglement metrics provide quantitative measures of the alignment between latent dimensions and underlying generative factors, as well as the degree of independence among latent variables. These metrics can be broadly categorised into two groups: \emph{dimension-based} and \emph{factor-based} metrics.

\textbf{\textit{(1) Dimension-based metrics}} evaluate disentanglement by analysing how individual latent dimensions respond to variations in generative factors. Common examples include $z_{\text{min-var}}$ error~\cite{kim2018disentangling}, \textit{Modularity}~\cite{ridgeway2018learning}, and the \textit{Disentanglement Score}~\cite{eastwood2018framework}. These metrics assess whether a given latent dimension remains invariant to changes in unrelated factors, indicating that it encodes a distinct and isolated feature.

\textbf{\textit{(2) Factor-based metrics}} assess disentanglement from the perspective of individual generative factors by measuring how latent dimensions respond when a single factor varies. Representative metrics include the \textit{Mutual Information Gap} (MIG)~\cite{chen2018isolating}, \textit{Completeness} and \textit{Informativeness}~\cite{eastwood2018framework}, and \textit{Explicitness}~\cite{ridgeway2018learning}.

We summarise the definitions of the commonly used metrics below:
\begin{enumerate}
    \item $z_{diff}~accuracy$~\cite{higgins2016beta}: The accuracy of a predictor for $p(y|z^b_{diff})$, where $z^b_{diff}$ is the absolute linear difference between the inferred latent representations for a batch $B$ of latent vectors, written as a percentage value. Higher values imply better disentanglement.

    \item $z_{min\_var}~error$~\cite{kim2018disentangling}: For a chosen factor k, data is generated with this factor fixed but all other factors varying randomly; their representations are obtained, with each dimension normalised by its empirical standard deviation over the full data (or a large enough random subset); the empirical variance is taken for each dimension of these normalised representations. Then the index of the dimension with the lowest variance and the target index k provide one training input/output example for the classifier. Thus, if the representation is perfectly disentangled, the empirical variance in the dimension corresponding to the fixed factor will be 0. The representations are normalised so that the arg min
    is invariant to rescaling of the representations in each dimension. Since both inputs and outputs lie in a discrete space, the optimal classifier is the majority-vote classifier, and the metric is the error rate of the classifier. Lower values imply better disentanglement.

    \item Mutual Information Gap ($MIG$)~\cite{chen2018isolating}: The difference between the top two latent variables with the highest mutual information. Empirical mutual information between a latent representation $z_j$ and a ground truth factor $v_k$, is estimated using the joint distribution defined by  $q(z_j, v_k) = \sum_{n=1}^{N}{p(v_k)p(n|v_k)q(z_j|n)}$. A higher mutual information implies that $z_j$ contains a more information about $v_k$, and the mutual information is maximal if there exists a deterministic, invertible relationship between $z_j$ and $v_k$. MIG values are in the interval [0, 1], with higher values implying better disentanglement.

    \item \textit{Modularity}~\cite{ridgeway2018learning}: The deviation from an ideally modular case of latent representation. If latent vector dimension $i$ is ideally modular, it will have high mutual information with a single factor and zero mutual information with all other factors. A deviation $\delta_i$ of 0 indicates perfect modularity and 1 indicates that this dimension has equal mutual information with every factor. Thus, $1 - \delta_i$ is used as a modularity score for vector dimension i and the mean of $1 - \delta_i$ over $i$ as the modularity score for the overall representation. Higher values imply better disentanglement.

    \item \textit{Explicitness}~\cite{ridgeway2018learning}: Mean of the ROC area-under-the-curve ($AUC_{jk}$) of a one-versus-rest logistic-regression classifier that takes the latent vectors as input and has factor values as targets, over a factor index $j$ and an index $k$ on values of factor $j$. Represents the coverage of the representation, in other words, how well each factor is represented. Higher values imply better disentanglement.

    \item \textit{Disentanglement Score}~\cite{eastwood2018framework}: The degree to which a representation factorises or disentangles the underlying factors of variation, with each variable (or dimension) capturing at most one generative factor. It is computed as a weighted average of a disentanglement score $D_i = (1 -  H_K(P_i.))$ for each latent dimension variable $c_i$, on the relevance of each $c_i$, where $H_K(P_i.)$ denotes the entropy and $P_{ij}$ denotes the 'probability' of $c_i$ being important for predicting $z_j$. If $c_i$ is important for predicting a single generative factor, the score will be 1. If $c_i$ is equally important for predicting all generative factors, the score will be 0. Higher values imply better disentanglement.

    \item \textit{Completeness Score}~\cite{eastwood2018framework}: The degree to which each underlying factor is captured by a single latent dimension variable. For a given $z_j$ it is given by $C_j = (1 - H_D(\tilde{P}.j))$, where $H_D(\tilde{P}.j) = -\sum_{d=0}^{D - 1}{\tilde{P}_{dj} log_D \tilde{P}_{ij}}$ denotes the entropy of the $\tilde{P}.j$ distribution. If a single latent dimension variable contributes to $z_j$'s prediction, the score will be 1 (complete). If all code variables contribute equally to $z_j$'s prediction, the score will be 0 (maximally over-complete). Higher values imply better disentanglement.

    \item \textit{Informativeness Score}~\cite{eastwood2018framework}: The amount of information that a representation captures about the underlying factors of variation. Given a latent representation $c$, It is quantified for each generative factor $z_j$ by the prediction error $E(z_j , \hat{z}_j)$ (averaged over the dataset), where $E$ is an appropriate error function and $\hat{z}_j = f_j(c)$. Lower values imply better disentanglement.
\end{enumerate}

\end{document}